\documentclass[preprint,review,12pt]{elsarticle}

\usepackage{lineno,hyperref}
\usepackage{amsmath}
\usepackage{prettyref}
\usepackage{subfigure}
\usepackage{float}
\usepackage{booktabs}

\hypersetup{hyperindex=false, colorlinks=true, linkcolor=blue, citecolor=blue, urlcolor=blue}
\newrefformat{fig}{Fig.~\ref{#1}}
\newrefformat{tbl}{Table~\ref{#1}}
\newrefformat{sec}{Section~\ref{#1}}
\newrefformat{eq}{Eq.~\ref{#1}}
\newrefformat{alg}{Algorithm?~\ref{#1}}

\makeatletter
\def\ps@pprintTitle{%
   \let\@oddhead\@empty
   \let\@evenhead\@empty
   \def\@oddfoot{\reset@font\hfil\thepage\hfil}
   \let\@evenfoot\@oddfoot
}
\makeatother

\biboptions{authoryear}

\begin{document}


\begin{frontmatter}

\title{A region-growing approach for automatic outcrop fracture extraction from a three-dimensional point cloud}

\author[zju]{Xin Wang}
\ead{ericrussell@zju.edu.cn}

\author[zju]{Lejun Zou}

\author[zju]{Xiaohua Shen}
\ead{shenxh@zju.edu.cn}

\author[zju]{Yupeng Ren}

\author[zju]{Yi Qin}

\address[zju]{Department of Earth Sciences, Zhejiang University Yuquan Campus, 38 Zheda Road, Hangzhou 310027, China}


\begin{abstract}

Conventional manual surveys of rock mass fractures usually require large amounts of time and labor; yet, they provide a relatively small set of data that cannot be considered representative of the study region. Terrestrial laser scanners are increasingly used for fracture surveys because they can efficiently acquire large area, high-resolution, three-dimensional (3D) point clouds from outcrops. However, extracting fractures and other planar surfaces from 3D outcrop point clouds is still a challenging task. No method has been reported that can be used to automatically extract the full extent of every individual fracture from a 3D outcrop point cloud. In this study, we propose a method using a region-growing approach to address this problem; the method also estimates the orientation of each fracture. In this method, criteria based on the local surface normal and curvature of the point cloud are used to initiate and control the growth of the fracture region. In tests using outcrop point cloud data, the proposed method identified and extracted the full extent of individual fractures with high accuracy. Compared with manually acquired field survey data, our method obtained better-quality fracture data, thereby demonstrating the high potential utility of the proposed method.

\end{abstract}


\end{frontmatter}


\section{Introduction}

The manual surveying of fractures and other planar rock mass surfaces is one of the most fundamental but time-consuming activities performed by field geologists. The surveyed fracture data usually comprise the fracture location, orientation, and surface roughness, which can support models and/or hypotheses in various applications (e.g., structural and geomechanical analysis, flow modeling, reservoir characterization, and engineering rock mass classification). These surveys are conventionally performed in situ with standard fieldwork instruments, such as a handheld compass, clinometer, and possibly a digital camera to record the fracture locations. However, the development of remote sensors (e.g., LiDAR-based scanners) and their availability as research equipment have prompted geoscientists to develop new methods that improve the analysis, avoid access problems, reduce time and labor, and result in a more representative dataset. The terrestrial laser scanner (TLS) is one of the most widely used instruments in Earth science applications, and it is very useful for acquiring high-quality, high-resolution, three-dimensional (3D) point clouds from outcrops \citep{Xu00,Bellian05,McCaffrey05,Olariu08,Jones09,Mah11,Wilson11,Pearce11,Mah13}. In addition, the GPS receiver module in a typical TLS allows the point cloud, a set of points in a 3D coordinate system, to be transformed into different geographic coordinate systems, so the data can be processed and used for different purposes, such as topographic feature extraction and orientation estimation for planar surfaces.

During the last few years, because of the widespread use of TLS in Earth science applications, there has been a growing need for point cloud processing methods to perform analyses and interpretation. The extraction of fractures and other planar surfaces from 3D outcrop point clouds has been the focus of much research by the geological community because fracture data have a wide range of applications. Many semi-automatic and automatic methods have been developed in the last 10 years, and certain algorithms can be used to extract (or segment) the points of fracture surfaces from the point cloud. \cite{Slob05} and \cite{Lato09} derived triangulated irregular networks from the point cloud and then grouped neighboring polygons with a similar orientation to obtain planar features. \cite{Roncella05}, \cite{Voyat06}, and \cite{Ferrero09} used random sample consensus (RANSAC) algorithm-based methods to segment point clouds into subsets, each of which comprise points that belong to the same discontinuity surface. Recently, other methods have also been proposed based on k-means clustering \citep{Olariu08}, moving sampling cube \citep{Gigli11}, point attributes \citep{Garcia11}, neighboring points coplanarity testing \citep{riquelme14}, and principal component analysis (PCA) \citep{Gomes16}. However, using these methods, either the full extent of the individual fracture surface is not extracted \citep{ Slob05, Olariu08,Lato09} or human supervision is required \citep{ Ferrero09, Gigli11, Garcia11, riquelme14}. The automatic method proposed by \cite{ Gomes16} splits the point cloud into four subsets (quadrants) iteratively to detect planar structures, but the full extent of the individual fracture surface was not fully extracted and was detected as several planar structures in most cases.

In this study, we propose an algorithm using a region-growing approach for the automatic extraction of the full extent of individual outcrop fractures from point clouds and for estimating their orientation. The main novel feature of this algorithm is the application of region growing to the extraction of outcrop fractures from point clouds. Instead of growing the region locally without a global view of the fracture surface, we use a seed point selection criterion to consider the overall fracture occurrence, as well as criteria for determining the initial seed point and controlling the growth of the region. The region-growing concept is simple, and by using carefully designed criteria, our algorithm can extract the full extent of every individual fracture in an automatic and robust manner.

\section{Study area and database}
\label{sec:study_area_and_point_cloud_acquisition}

\begin{figure}[H]
  \centering
  \includegraphics[width=0.56\textwidth]{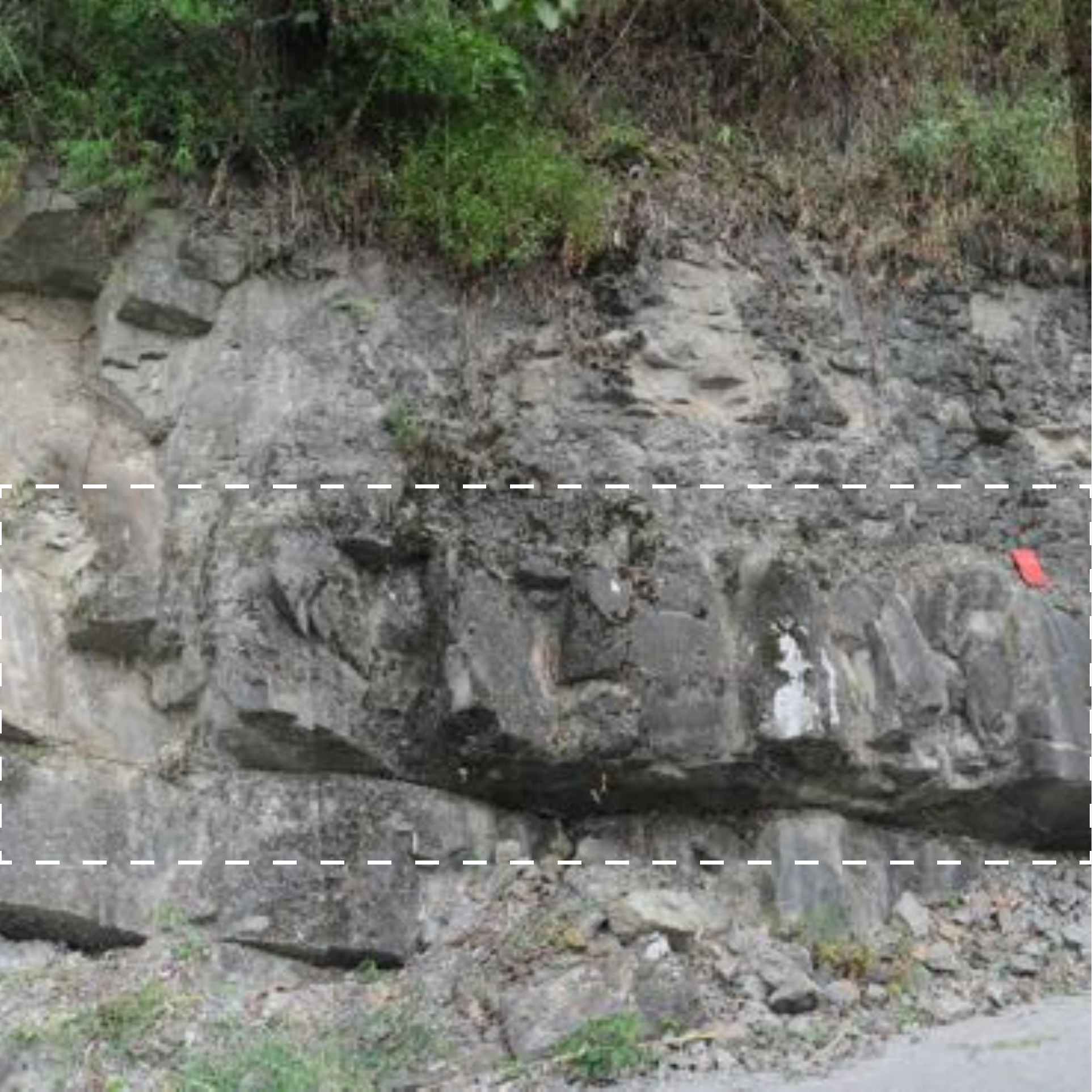}
  \caption{Picture of the study site: a road-cut rock slope with thin to thick layered sandstone. The white rectangle delineates the part used to test our algorithm. See the red notebook for scale.}
  \label{fig:figure_1}
\end{figure}

The study site is a road-cut rock slope located along a country road in Nanbaoxiang, Chengdu, Sichuan Province, China (N 30$^\circ$ 24$'$ 25.35$''$, E 103$^\circ$ 11$'$ 8.34$''$) (\prettyref{fig:figure_1}). The rock slope mainly comprises thin to thick layered sandstone, and the area of the study outcrop is about 30 m$^2$. A RIEGL VZ-1000 terrestrial laser scanning system (mainly comprising a 3D laser scanner, digital camera, and GPS receiver) was used to perform a high-resolution LiDAR scan of the rock slope. This TLS system uses the time-of-flight technique, which utilizes the emission and return time of highly collimated electromagnetic radiation to calculate the distance from the instrument's optical center to a reflecting target surface \citep{Baltsavias99}. An outcrop 3D point cloud was acquired with an average point spacing of $<$ 1 cm; there were about 21 million points. To test the proposed region-growing-based algorithm, we selected the central part of the point cloud (delineated by the white rectangle in \prettyref{fig:figure_1}), where less vegetation and fallen stone were present. Conventional measurements of fracture surface orientations using a handheld compass were also performed on the rock slope to compare with the results obtained by the proposed algorithm, and 65 orientation measurements (dip direction and dip angle) were acquired from fracture faces distributed over the outcrop.

\section{Methodology}

Region growing is an element-based segmentation method, which has many advantages compared with other methods. The concept of region growing is simple; only a small number of seed points and a few criteria are required to grow the region. Region growing can correctly segment regions that share the same defined properties. The seed points and the criteria can be selected freely to suit different applications. A drawback of region growing is that it lacks a global view of the problem; however, this drawback can be addressed by selecting criteria that consider the global view of the problem during the region growing, and this we do in our proposed algorithm.

The proposed region-growing-based algorithm works directly with the LiDAR point cloud instead of using an interpolated 2.5D mesh surface. Many detailed geometrical features extracted from the point cloud of the outcrop surface can be used for the segmentation, but the proposed algorithm uses mainly the local surface normal and curvature.

The proposed method comprises three main steps, as follows. First step: Local surface normal and curvature estimation, which involves a nearest-neighbor search as well as the estimation of the least-squares fitting plane and the curvature of the neighboring points. This task is described in \prettyref{sec:normal_and_curvature_estimation}. Second step: Region growing, which extracts the fracture face by using criteria based on the local surface normal and curvature to select the seed points and to control the growth. This step is explained in \prettyref{sec:region_growing}. Third step: Fracture orientation estimation, which employs a patch of the point cloud after its growth is complete. This part is described in \prettyref{sec:orientation_estimation}.

\subsection{Local surface normal and curvature estimation}
\label{sec:normal_and_curvature_estimation}

One or more properties of each point in the point cloud are required for region-growing segmentation, i.e., segmenting the regions that comprise points with similar defined properties. The local surface normal and the local surface curvature are two basic properties that can be used to define planar surfaces such as fractures.

To estimate the local surface normal and curvature for each point, its neighboring points, which together form the local topography, are needed. We refer to a point cloud as $P$, a collection of 3D points $\boldsymbol{p}_i = \{x_i, y_i, z_i\} \in P$. Let $\boldsymbol{p}_\mathrm{q}$ be the query point in the problem of estimating the local surface normal and curvature, and let $P^k$ be the K-nearest neighbors of $\boldsymbol{p}_\mathrm{q}$, in which $k$ is chosen by the user to find the $k$ nearest neighbors according to their Euclidean distance to $\boldsymbol{p}_\mathrm{q}$.

The method we use for estimating the local surface normal is based on least-squares plane fitting with $P^k$, as proposed by \cite{Berkmann94}. The least-squares plane fitting method is based on PCA. The local surface normal $\boldsymbol{n}_\mathrm{q}$ of point $\boldsymbol{p}_\mathrm{q}$ is obtained by analyzing the eigenvalues and eigenvectors of $P^k$'s covariance matrix $C = \frac{1}{k}\sum_{i=1}^k (\boldsymbol{p}_i - \bar{\boldsymbol{p}}) \cdot (\boldsymbol{p}_i - \bar{\boldsymbol{p}})^\mathrm{T}$, where $\boldsymbol{p}_i \in P^k$ and $\bar{\boldsymbol{p}} = \frac{1}{k} \cdot \sum_{i=1}^k \boldsymbol{p}_i$. If we let $\lambda_0$, $\lambda_1$, and $\lambda_2$ be the eigenvalues of $C$ that satisfy $0 \leq \lambda_0 \leq \lambda_1 \leq \lambda_2$ and if $\boldsymbol{v}_0$ is the corresponding eigenvector of $\lambda_0$, then
\begin{linenomath}
\begin{equation}
\boldsymbol{n}_\mathrm{q} = \begin{cases}
\boldsymbol{v}_0 & \text{if } \boldsymbol{v}_0 \cdot (\boldsymbol{v}_\mathrm{p} - \boldsymbol{p}_\mathrm{q}) > 0\\
-\boldsymbol{v}_0 & \text{if } \boldsymbol{v}_0 \cdot (\boldsymbol{v}_\mathrm{p} - \boldsymbol{p}_\mathrm{q}) < 0 ,
\end{cases}
\label{eq:local_surface_normal}
\end{equation}
\end{linenomath}
where $\boldsymbol{v}_\mathrm{p}$ is the viewpoint from which the point cloud is acquired.

The method used to estimate the local surface curvature was proposed by \cite{Pauly02}. This method can estimate the curvature directly from the eigenvalues of $P^k$'s covariance matrix $C$ without needing to first create a surface from the point cloud. The local surface curvature $\sigma_\mathrm{q}$ of $\boldsymbol{p}_\mathrm{q}$ is estimated as follows.
\begin{linenomath}
\begin{equation}
\sigma_\mathrm{q} = \frac{\lambda_0}{\lambda_0 + \lambda_1 + \lambda_2}
\label{eq:local_surface_curvature}
\end{equation}
\end{linenomath}

\subsection{Region growing}
\label{sec:region_growing}

Let $P_\mathrm{r} \subset P$ be the set of points that have not yet been assigned to any fracture regions. For each fracture region, the initial seed point that starts this region's growth is selected from $P_\mathrm{r}$, and the point $\boldsymbol{p}_\mathrm{min} \in P_\mathrm{r}$ with the minimum curvature is selected as a reasonable initial seed point for planar surfaces such as fractures.

Next, the criterion that controls the growth from the seed points to their neighboring points is defined as the local surface normal deviation threshold $\theta_\mathrm{th}$ given by the user. For the neighboring point $\boldsymbol{p}_i$, the local surface normal is $\boldsymbol{n}_i$, and the seed point's local surface normal is $\boldsymbol{n}_\mathrm{s}$, so $\boldsymbol{p}_i$ is added to the current region if $\cos^{-1}(\frac{\boldsymbol{n}_\mathrm{s} \cdot \boldsymbol{n}_i}{\lVert \boldsymbol{n}_\mathrm{s} \rVert \lVert \boldsymbol{n}_i \rVert}) < \theta_\mathrm{th}$, i.e., if the angle between $\boldsymbol{n}_\mathrm{s}$ and $\boldsymbol{n}_i$ is less than $\theta_\mathrm{th}$.

The new seed points are then selected from the newly added points. The criterion defined as the transmission error threshold $t_\mathrm{th}$, which is also given by the user, determines whether the newly added $\boldsymbol{p}_i$ is selected as a new seed point. For the newly added $\boldsymbol{p}_i$, the local surface normal is $\boldsymbol{n}_i$ and the initial seed point $\boldsymbol{p}_\mathrm{min}$'s local surface normal is $\boldsymbol{n}_\mathrm{min}$, so the newly added $\boldsymbol{p}_i$ is selected as a new seed point if $\cos^{-1}(\frac{\boldsymbol{n}_\mathrm{min} \cdot \boldsymbol{n}_i}{\lVert \boldsymbol{n}_\mathrm{min} \rVert \lVert \boldsymbol{n}_i \rVert}) < t_\mathrm{th}$, i.e., if the angle between $\boldsymbol{n}_\mathrm{min}$ and $\boldsymbol{n}_i$ is less than $t_\mathrm{th}$. Therefore, the overall occurrence of the fracture, which is represented by $\boldsymbol{n}_\mathrm{min}$, serves to control the growth of the region instead of its being allowed to grow blindly.

The region's growth from the newly selected seed points and the selection of new seed points are then performed iteratively until no new seed point can be selected and the region's growth is complete. The growth of the other regions is completed for those remaining in the point cloud until all of the points in the point cloud have been processed.

\begin{figure}[H]
  \centering
  \includegraphics[width=0.7\textwidth]{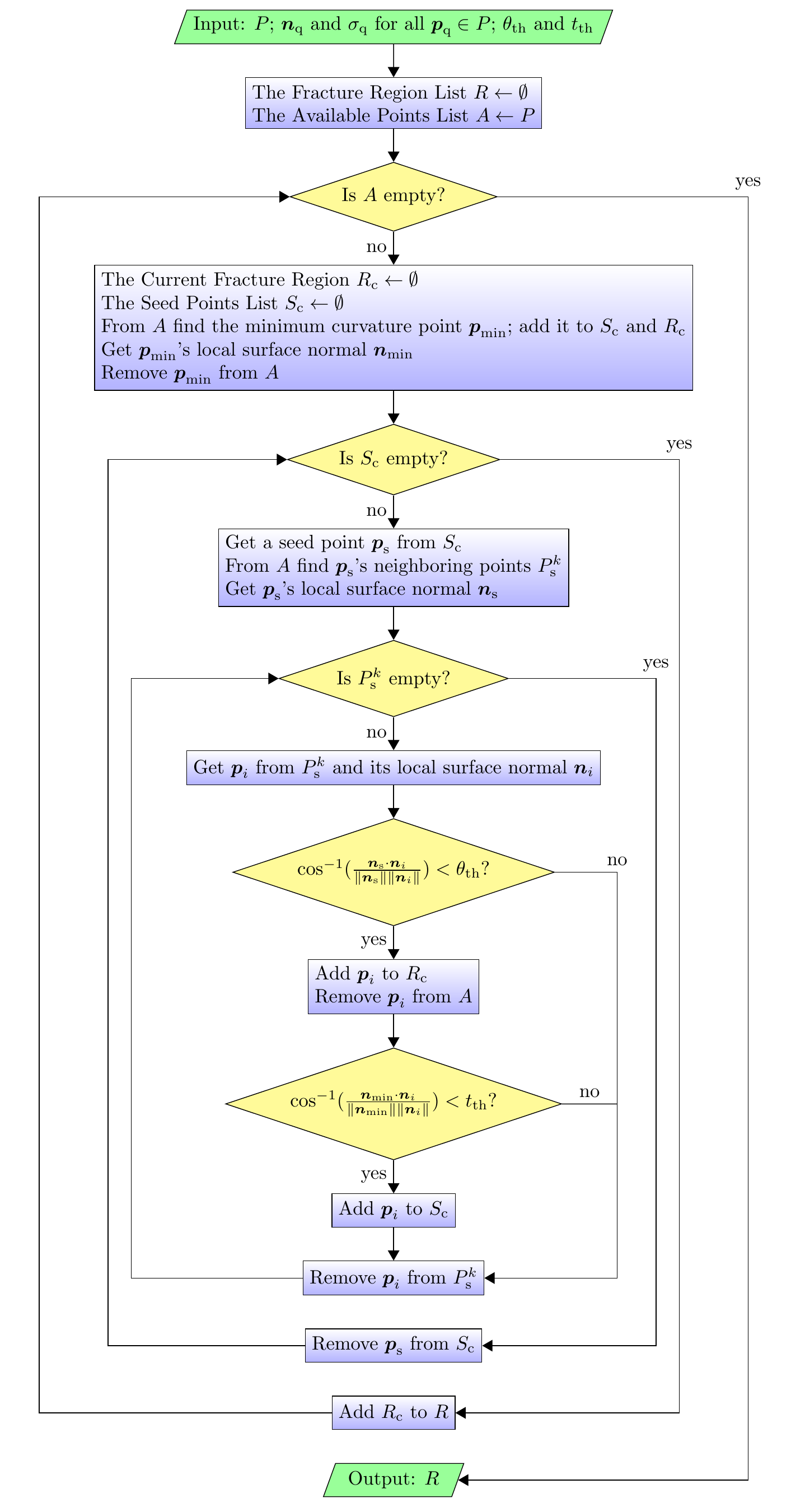}
  \caption{Flowchart illustrating the region-growing algorithm used by the proposed method.}
  \label{fig:flowchart}
\end{figure}

The local surface normal deviation threshold $\theta_\mathrm{th}$ and the transmission error threshold $t_\mathrm{th}$ that yield the best segmentation result are related to the fracture surface's geometrical nature and the weathering condition of the outcrop. According to their definitions and the functions described above, $\theta_\mathrm{th}$ is related to the local roughness, whereas $t_\mathrm{th}$ is related to the overall flatness of the fracture. For example, if a flat fracture has a rough local surface, then $\theta_\mathrm{th}$ should be sufficiently large to allow small protrusions and dents in the fracture region. If the uneven fracture has a smooth local surface, such as weathered fracture surfaces, then $t_\mathrm{th}$ should be sufficiently large to allow the uneven fracture to grow into one region. The flow chart for the region-growing step of the proposed method is shown in \prettyref{fig:flowchart}.

\subsection{Fracture orientation estimation}
\label{sec:orientation_estimation}

For a patch of a point cloud that has completed its growth, many features of the extracted fracture can be estimated, such as the fracture orientation. The fracture orientation can be estimated as the normal $\langle n_x, n_y, n_z \rangle$ of the least-squares fitting plane in the point cloud patch that represents the fracture. The normal can also be transformed into the $\langle$~Dip~direction, Dip~$\rangle$, and the transformation may vary with the geographic coordinate systems employed for the point cloud. For example, if the y-axis of the coordinate system points north, the x-axis points east, and the z-axis points vertically up, then the Dip direction and Dip will be as follows.

\begin{equation}
\text{Dip direction} = \begin{cases}
0^\circ & \text{if } n_x = 0 \text{ \& } n_y \geq 0\\
180^\circ & \text{if } n_x = 0 \text{ \& } n_y < 0\\
90^\circ - \tan^{-1}(n_y/n_x) & \text{if } n_x > 0\\
270^\circ - \tan^{-1}(n_y/n_x) & \text{if } n_x < 0
\end{cases}
\label{eq:dip_direction}
\end{equation}

\begin{equation}
\text{Dip} = \begin{cases}
0^\circ & \text{if } n_x^2 + n_y^2 = 0\\
90^\circ - \tan^{-1}(\frac{\lvert n_z \rvert}{\sqrt{n_x^2 + n_y^2}}) & \text{if } n_x^2 + n_y^2 \neq 0
\end{cases}
\label{eq:dip}
\end{equation}

\section{Results and discussion}

The point cloud used to test our algorithm and the conventional measurements of the fracture surface orientations for the same outcrop were described in \prettyref{sec:study_area_and_point_cloud_acquisition}. The application of the proposed algorithm to the entire point cloud will provide a great number of planar regions with various dimensions, and this makes it difficult to show the detailed results obtained by the proposed algorithm. Therefore, before applying the proposed algorithm to the entire point cloud, we tested the algorithm with a portion of the point cloud, as described in \prettyref{sec:details_of_the_result}. The results obtained for the entire point cloud and comparisons with manual field survey results are discussed in \prettyref{sec:overall_result_and_its_comparison}. The performance of the proposed algorithm in terms of time consumption was tested using a set of outcrop point cloud data, and the number of planes detected from the same point cloud data with different configurations of $\theta_\mathrm{th}$ and $t_\mathrm{th}$ was also investigated. These are discussed in \prettyref{sec:performance_and_parameter_configuration}.

\subsection{Details of results for a portion of the point cloud}
\label{sec:details_of_the_result}

\begin{figure}[H]
  \centering
  \subfigure[]
  {\includegraphics[width=0.38\textwidth]{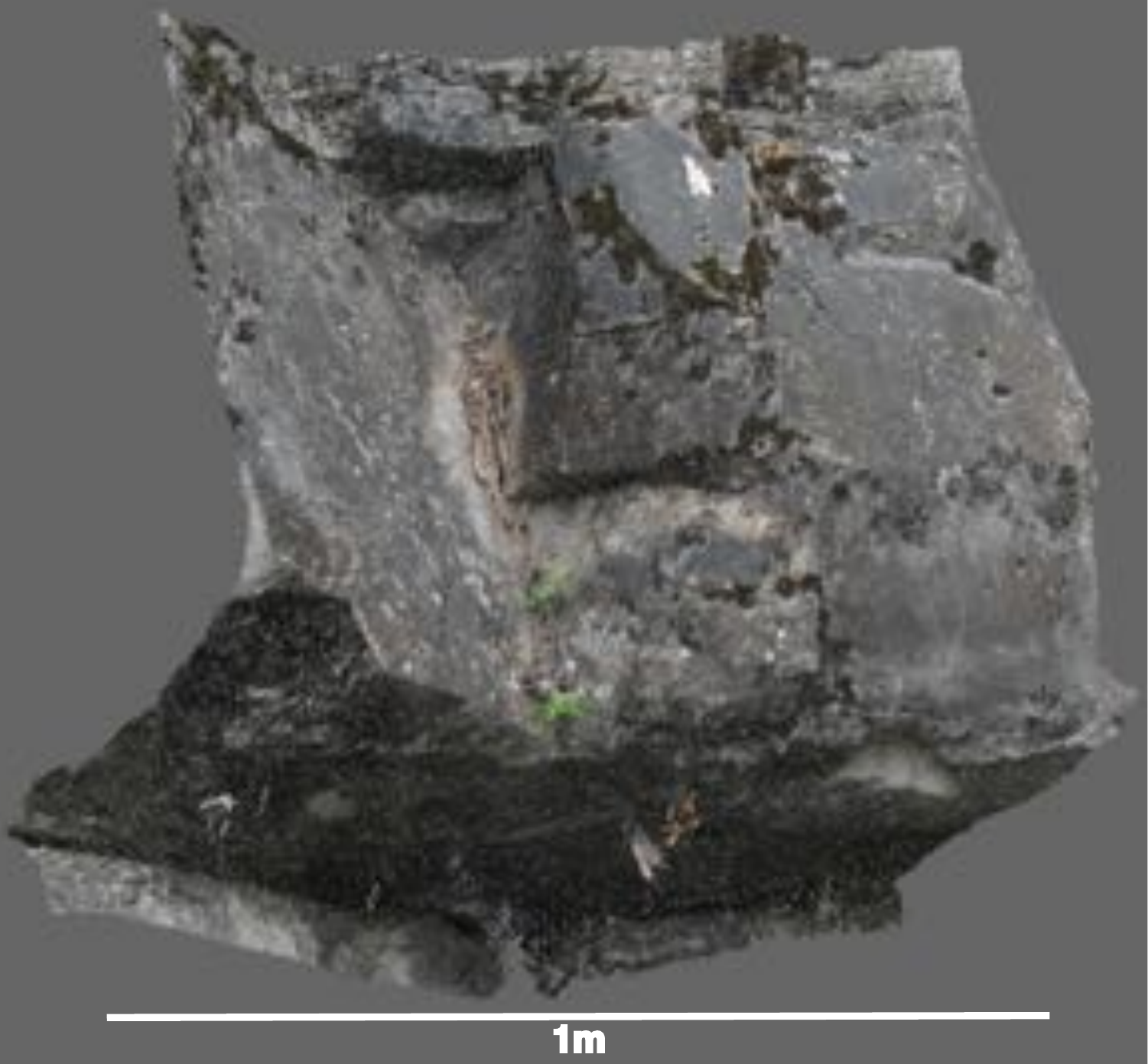}}
  \subfigure[]
  {\includegraphics[width=0.38\textwidth]{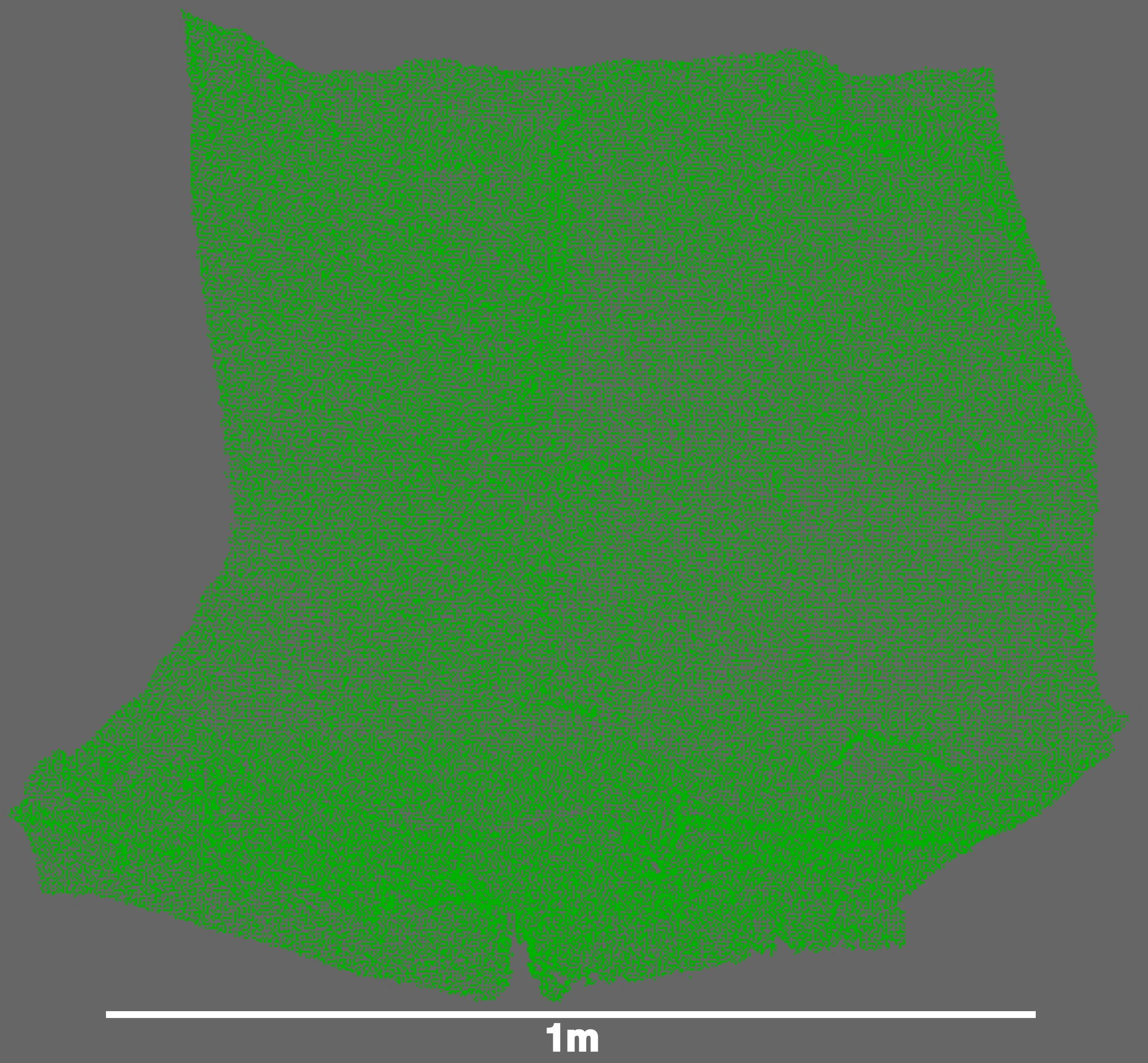}}
  \subfigure[]
  {\includegraphics[width=0.38\textwidth]{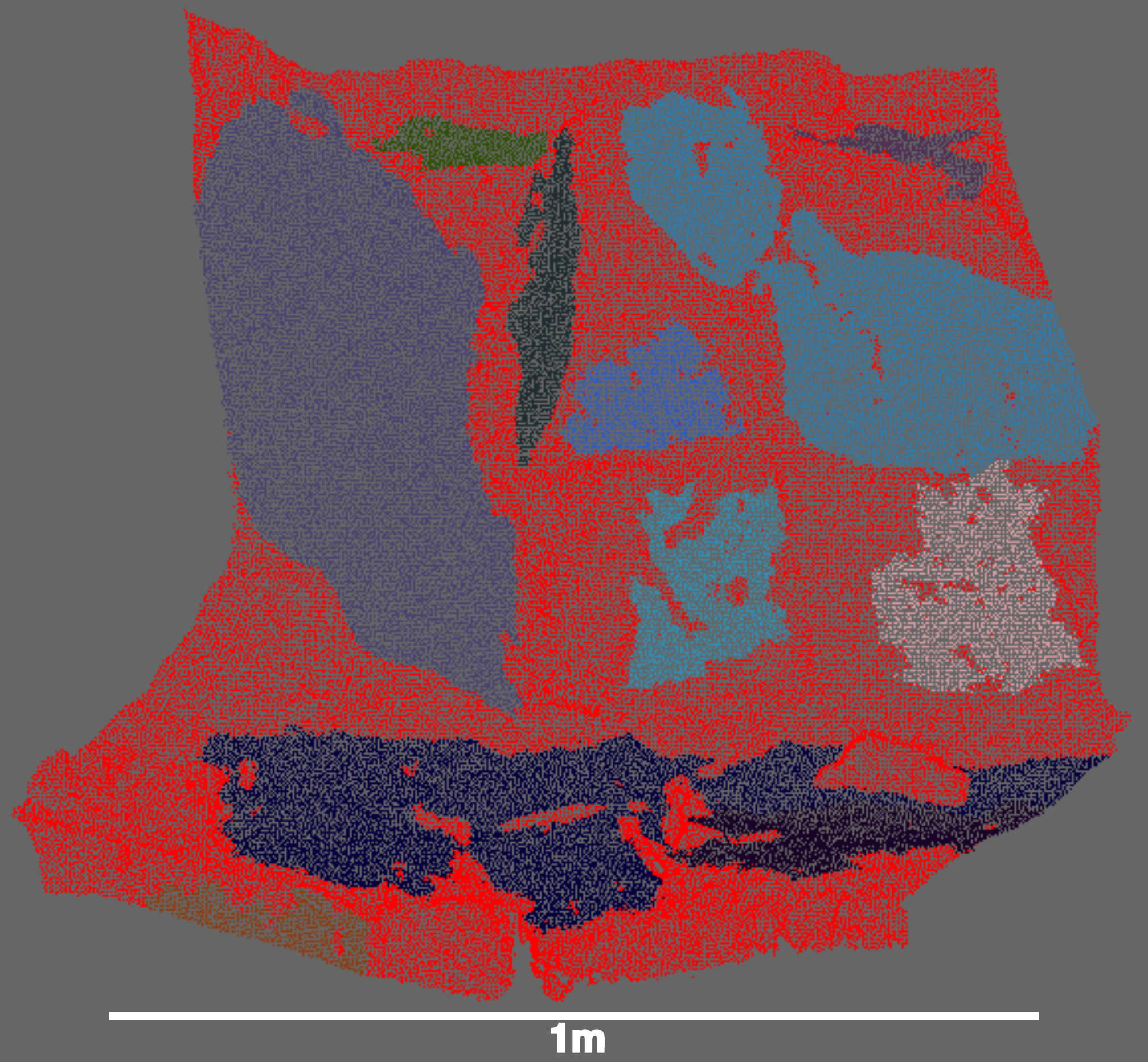}}
  \subfigure[]
  {\includegraphics[width=0.38\textwidth]{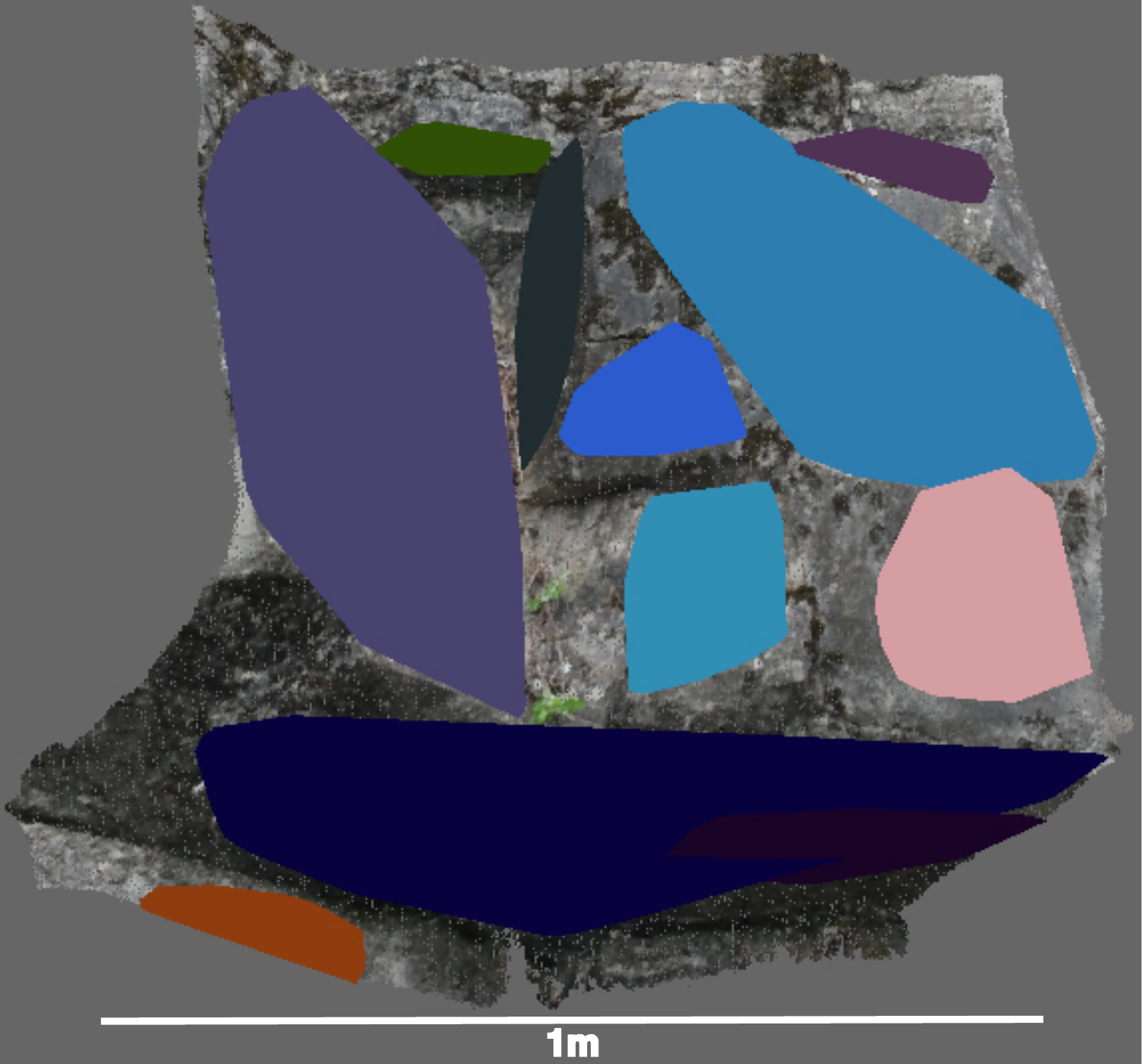}}
  \caption{Detailed results obtained from a portion of the point cloud. (a) A portion of the outcrop and (b) its point cloud. (c) Segmentation results obtained using the proposed algorithm. Different fracture regions are shown in various colors, and the non-fracture regions are shown in red. (d) Estimated fracture planes obtained from the segmentation results.}
  \label{fig:local_point_cloud_segmentation_result}
\end{figure}

A portion of the outcrop (\prettyref{fig:local_point_cloud_segmentation_result}a and its point cloud \prettyref{fig:local_point_cloud_segmentation_result}b) was processed using our method. The local surface normal deviation threshold $\theta_\mathrm{th}$ was set to 6$^\circ$, and the transmission error threshold $t_\mathrm{th}$ was set to 20$^\circ$, which were tuned to obtain the best results.  These settings for $\theta_\mathrm{th}$ and $t_\mathrm{th}$ can be applied to similar outcrop conditions. \prettyref{fig:local_point_cloud_segmentation_result}c is the result of the proposed algorithm; it shows the fracture regions having more than 800 points so that they can be conveniently compared with the manually identifiable fractures in \prettyref{fig:local_point_cloud_segmentation_result}a. The threshold value of 800 can be modified if smaller or larger fracture regions are required. Different fracture regions are indicated by different colors; the non-fracture regions are shown in red.

The results show that most of the fractures extracted by the proposed algorithm could be identified as real fractures in \prettyref{fig:local_point_cloud_segmentation_result}a and also that most of the major fractures identified in \prettyref{fig:local_point_cloud_segmentation_result}a were extracted by the proposed algorithm. In addition, the results show that unlike other existing methods, the proposed region-growing-based algorithm can extract the full extent of every individual fracture automatically.

The estimated fracture planes and the least-squares fitting planes for the extracted fracture regions are shown in \prettyref{fig:local_point_cloud_segmentation_result}d. The fracture orientations were estimated using these planes according to the methods described in \prettyref{sec:orientation_estimation}.

\subsection{Results for entire point cloud and comparison with manual field survey results}
\label{sec:overall_result_and_its_comparison}

\begin{figure}[H]
  \centering
  \subfigure[]
  {\includegraphics[width=0.98\textwidth]{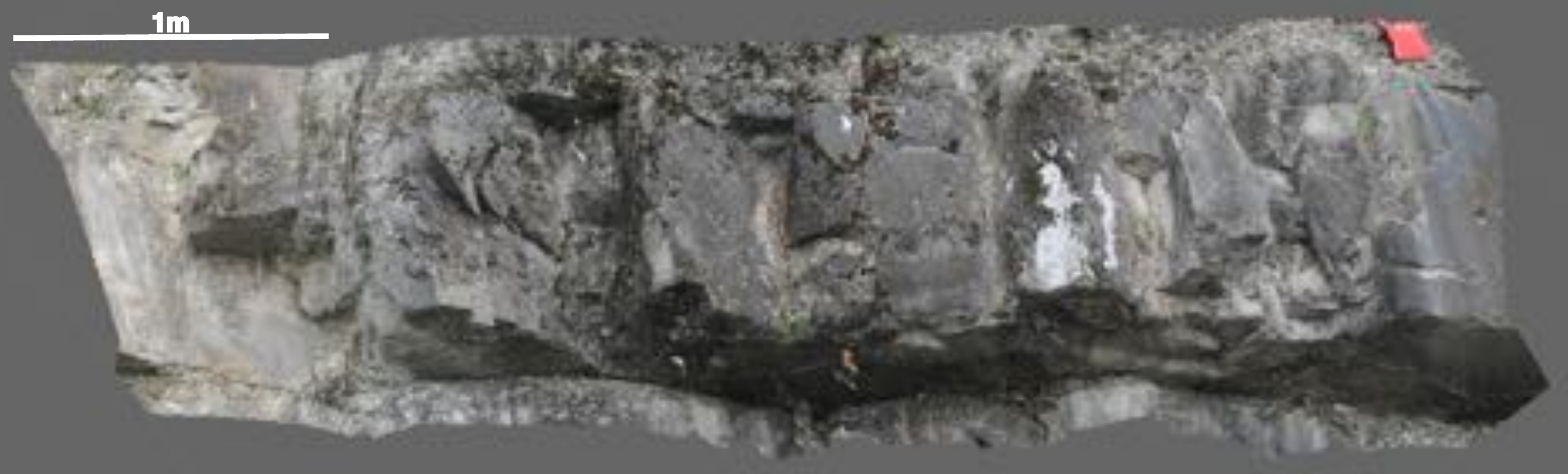}}
  \subfigure[]
  {\includegraphics[width=0.98\textwidth]{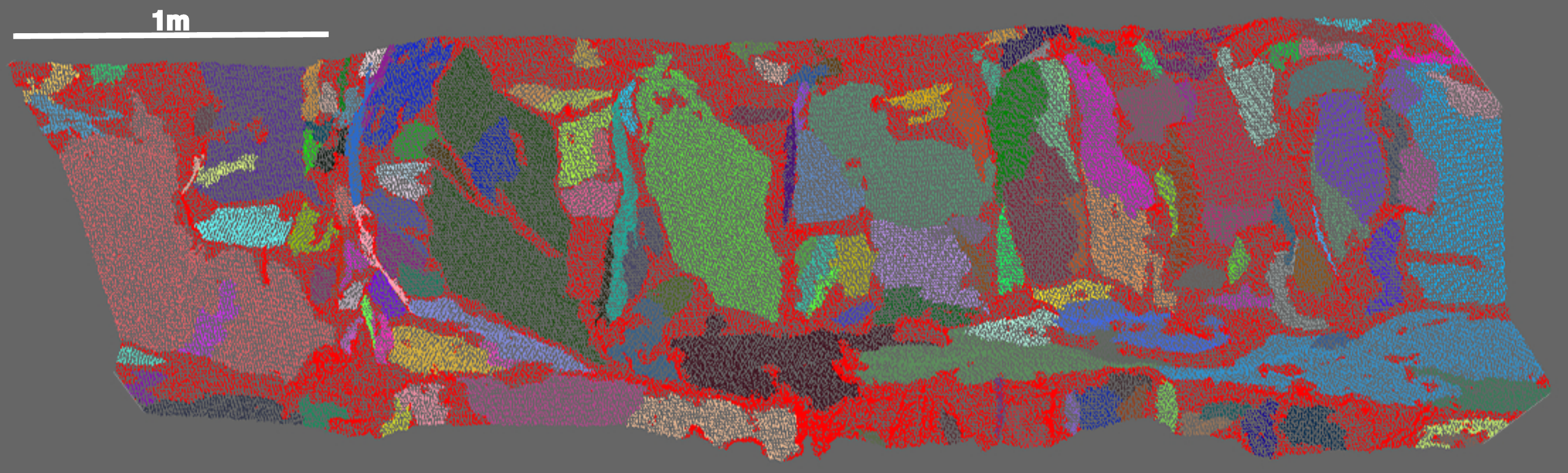}}
  \caption{(a) The outcrop used to test the proposed algorithm and (b) the segmentation results. Different fracture regions are shown by different colors, and the non-fracture regions are shown in red.}
  \label{fig:overall_point_cloud_segmentation_result}
\end{figure}

We applied the proposed method, with the local surface normal deviation threshold $\theta_\mathrm{th}$ set to 6$^\circ$ and the transmission error threshold $t_\mathrm{th}$ set to 20$^\circ$, to the entire point cloud (\prettyref{fig:overall_point_cloud_segmentation_result}a). The resulting fracture regions having more than 100 points are shown by different colors in \prettyref{fig:overall_point_cloud_segmentation_result}b, and the non-fracture regions are shown in red.

The results demonstrate that our proposed algorithm can extract many small fracture faces, cases for which conventional measurements cannot be obtained. Thus, besides accurately extracting fractures in the same manner as a conventional manual survey as demonstrated by the detailed results shown in \prettyref{sec:details_of_the_result}, our method may also provide additional information about fractures (particularly small fractures) that cannot be acquired from conventional manual surveys.

\begin{figure}[H]
  \centering
  \subfigure[]
  {\includegraphics[width=0.62\textwidth]{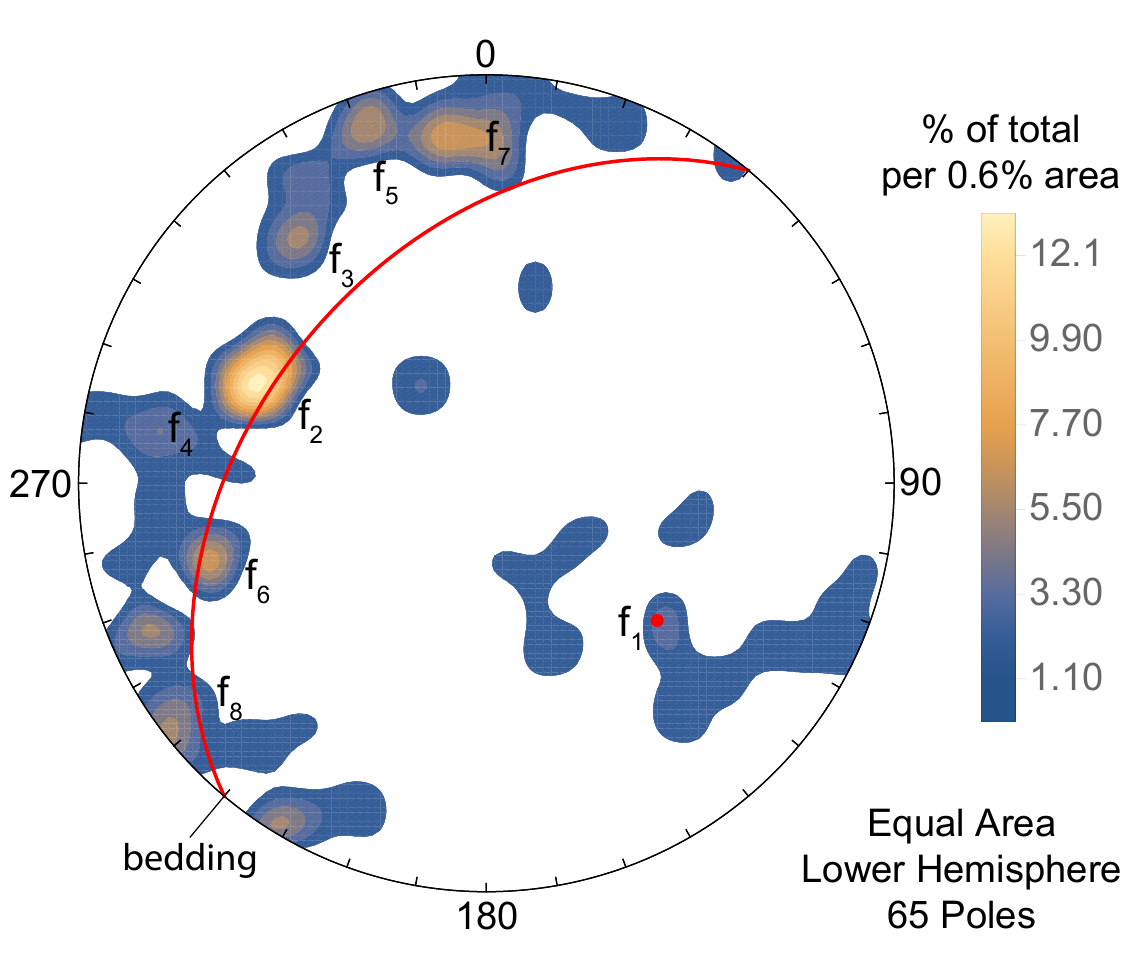}}
  \subfigure[]
  {\includegraphics[width=0.62\textwidth]{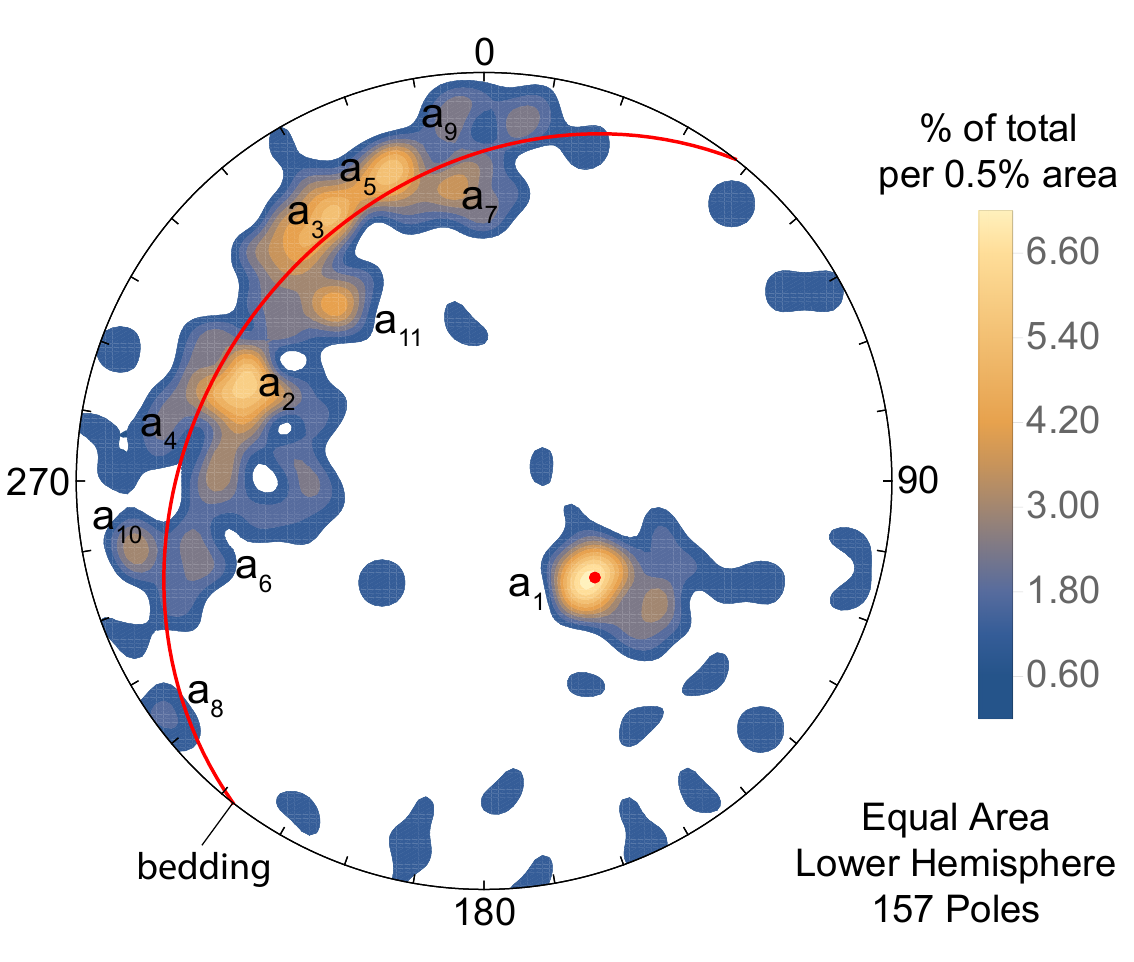}}
  \caption{Pole density plots for the fracture orientations obtained from (a) the manual field survey and (b) the results produced using the proposed algorithm. The bedding's pole and arc plot are each shown in red in both (a) and (b). $f_1,...,f_8$ and $a_1,...,a_8$ represent different sets of corresponding fractures.}
  \label{fig:field_contour_auto_contour}
\end{figure}

We extracted 157 fracture regions having more than 100 points and estimated their orientations. To compare the performance of the manual field survey and our proposed algorithm, we stereographically projected 65 orientations from the manual field survey and 157 orientations from the results obtained by the proposed algorithm and plotted the density of their poles (\prettyref{fig:field_contour_auto_contour}a and \prettyref{fig:field_contour_auto_contour}b, respectively).

A comparison of the pole density plots shows that the major clusters of poles representing different sets of fractures from the field manual survey, such as $f_1,...,f_8$ in \prettyref{fig:field_contour_auto_contour}a, can also be found in the results obtained using our algorithm ($a_1,...,a_8$ in \prettyref{fig:field_contour_auto_contour}b). The comparison also shows that the proposed algorithm has the advantage of locating clusters of fracture sets more accurately. For example, if many fractures are perpendicular to the bedding (as was found at the study site considered), then the stereographic plot poles of those fractures should be near the arc of the bedding, and obviously the proposed algorithm has better descriptions than the manual field survey. Thus, the red arc of bedding in \prettyref{fig:field_contour_auto_contour}b fits the distribution of poles better than the red arc of bedding in \prettyref{fig:field_contour_auto_contour}a, which indicates that our algorithm was better at locating the cluster of bedding and the other clusters of fracture sets. Therefore, in general, our algorithm was able to obtain fracture data whose quality was as good as or better than that from the manual field survey.

Furthermore, our algorithm provides additional information about fractures, which may be useful for the analysis. For example, clusters $a_9$, $a_{10}$ and $a_{11}$ in \prettyref{fig:field_contour_auto_contour}b, as well as the symmetries between $a_6$ and $a_7$, $a_9$ and $a_{10}$, $a_6$ and $a_{10}$, and $a_7$ and $a_9$, may be interesting information that merits further discussion and study.

The only disadvantage of the proposed algorithm may be the presence of some possible outlier clusters in the fracture sets (examples can be seen in \prettyref{fig:field_contour_auto_contour}b). The nature and removal of these outliers should be studied in future work.

\subsection{Performance and parameter configuration of the proposed algorithm}
\label{sec:performance_and_parameter_configuration}

A set of point cloud data (\prettyref{tbl:datasets_for_performance_testing}) was used to test the performance of our algorithm using a desktop computer with a CPU of 3.60 GHz and 4 GB RAM. The average point spacing in all the point cloud datasets used was 0.01 m, the dataset size ranged from 134,067 points to 1,096,948 points, and the outcrop areas ranged from 5.5 m$^2$ to 67.3 m$^2$. The same processing parameters were used for all the point cloud datasets to highlight the variation in the time consumption.

\begin{table}[H]
   \centering
   \caption{Point cloud datasets used to test the performance of the proposed region-growing-based algorithm} 
   \begin{tabular}{*{4}{c}} 
      \toprule
      Dataset  & Number of points & Average point spacing (m) & Area (m$^2$)\\
      \midrule
      1 & 134,067 & 0.01 & 5.5 \\
      2 & 323,562 & 0.01 & 14.5\\
      3 & 484,658 & 0.01 & 19.0\\
      4 & 684,866 & 0.01 & 33.7\\
      5 & 1,096,948 & 0.01 & 67.3\\
      \bottomrule
   \end{tabular}
   \label{tbl:datasets_for_performance_testing}
\end{table}

\begin{figure}[H]
  \centering
  \includegraphics[width=0.75\textwidth]{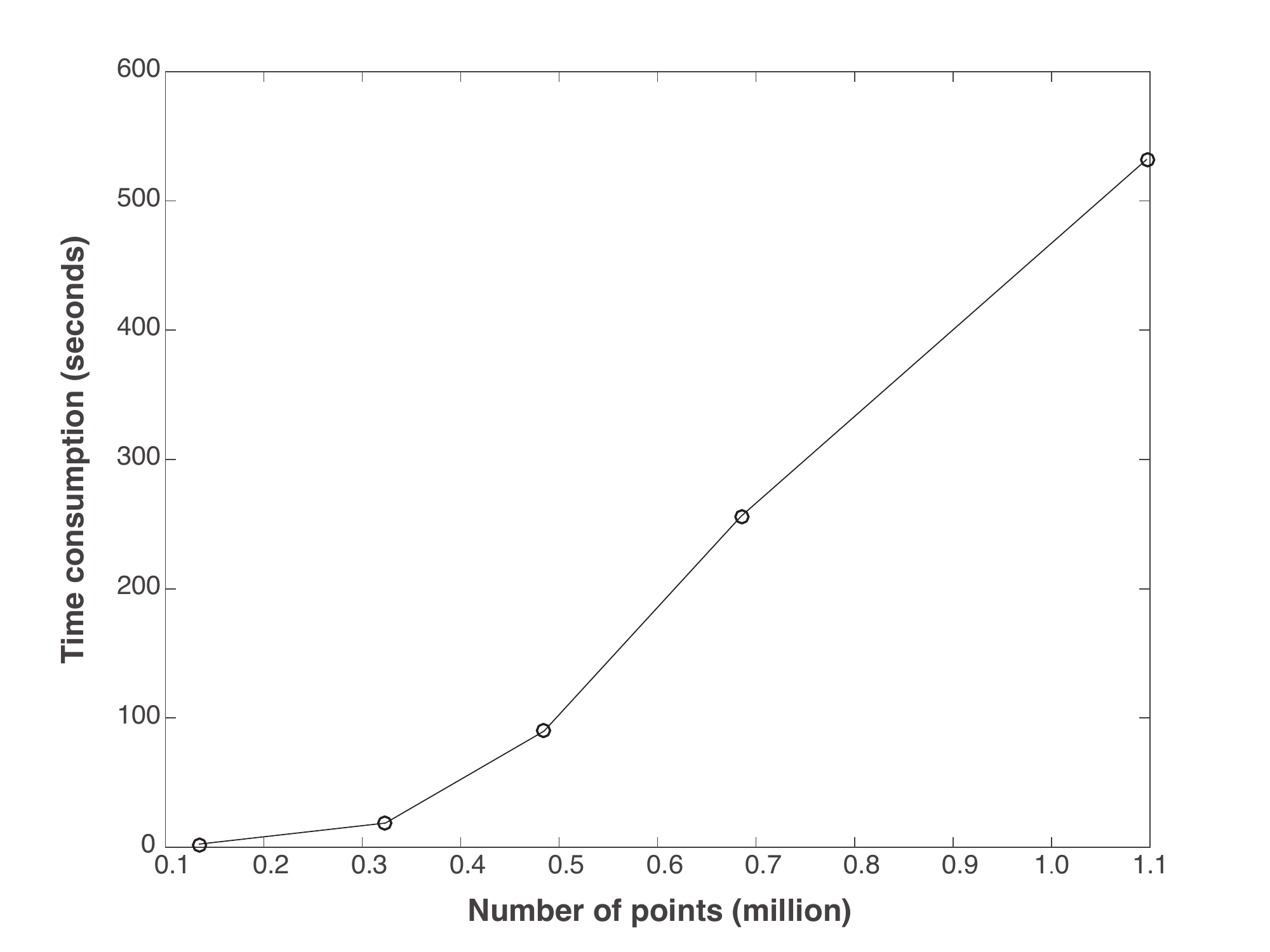}
  \caption{Time consumption of the proposed region-growing-based algorithm as point cloud size (number of points) increases.}
  \label{fig:performance}
\end{figure}

The performance testing result is shown in \prettyref{fig:performance}. The figure shows that there was a steep increase in time consumption as the point cloud size reached 1 million points, but the time consumption is still acceptable. Therefore, we conclude that our algorithm is suitable for point cloud datasets whose outcrop area $\le$ 70 m$^2$. For datasets larger than that, the computing power should be increased or the algorithm should be modified.

\begin{figure}[H]
  \centering
  \subfigure[]
  {\includegraphics[width=0.70\textwidth]{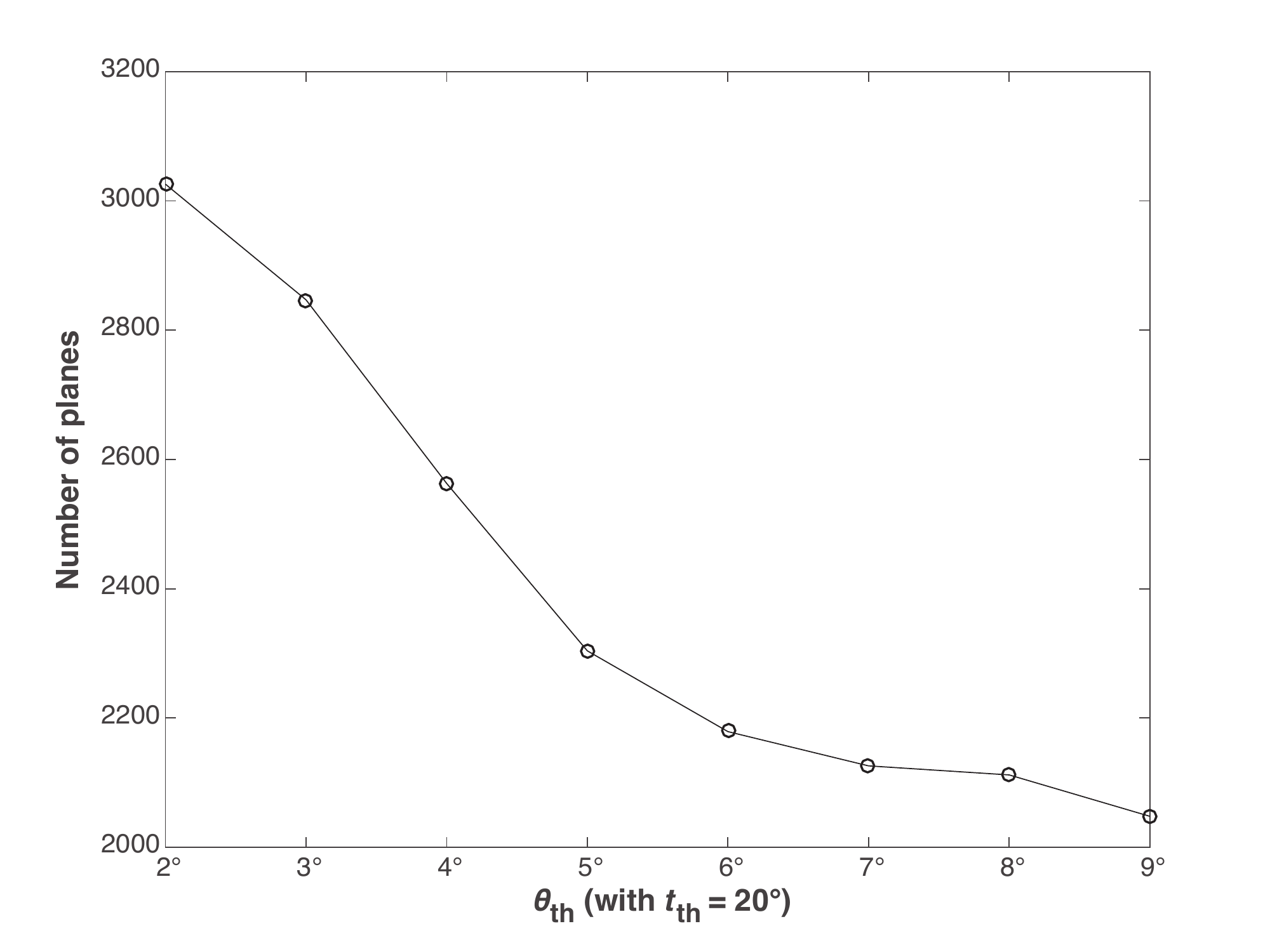}}
  \subfigure[]
  {\includegraphics[width=0.70\textwidth]{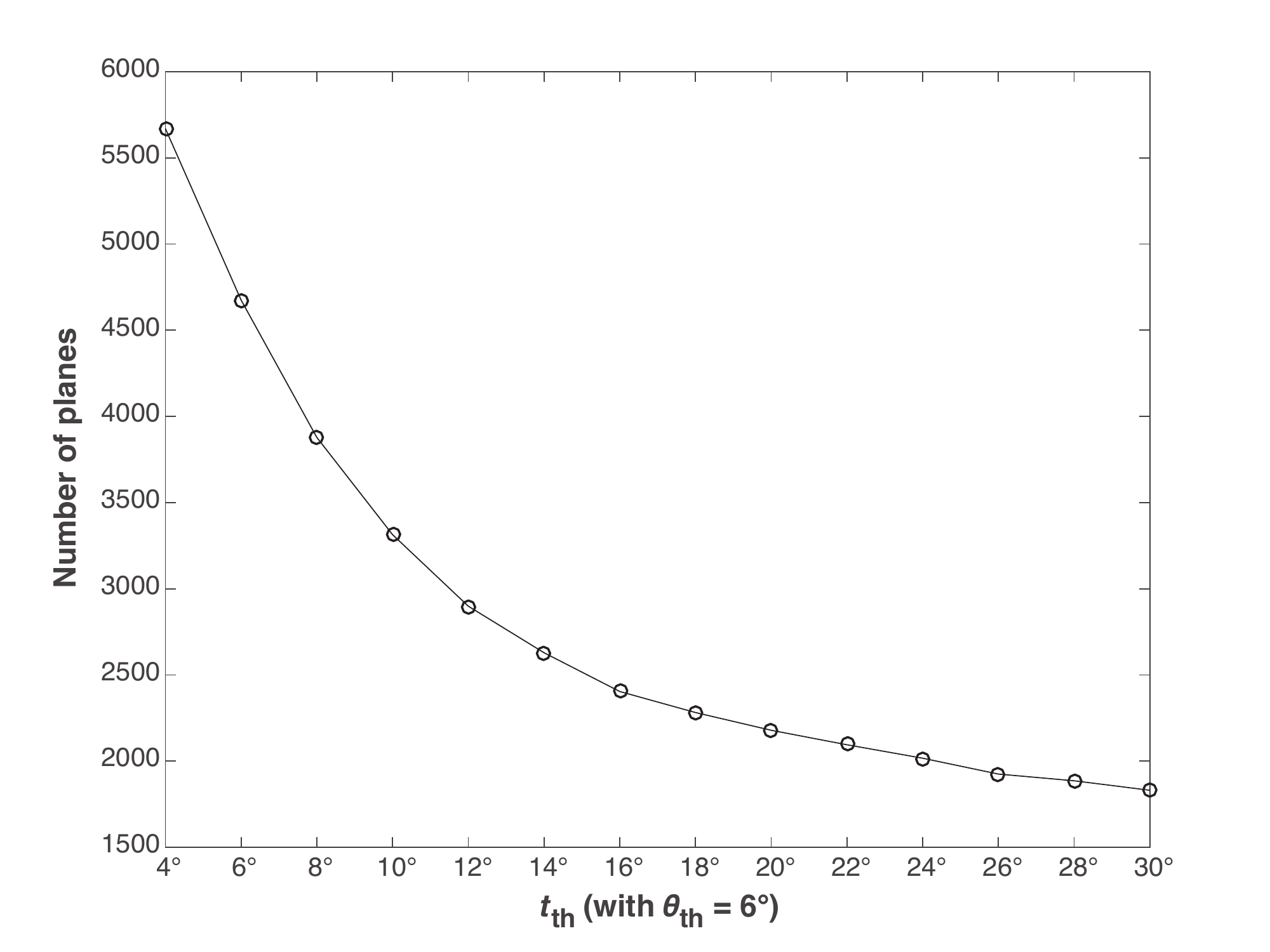}}
  \caption{Number of planes detected from the same point cloud under different configurations of (a) $\theta_\mathrm{th}$ and (b) $t_\mathrm{th}$.}
  \label{fig:p2theta_p2t}
\end{figure}

Using the same point cloud as the one shown in \prettyref{fig:overall_point_cloud_segmentation_result}a, we investigated the number of planes detected using different configurations of $\theta_\mathrm{th}$ and $t_\mathrm{th}$. The results are shown in \prettyref{fig:p2theta_p2t}, which shows that an increase in $\theta_\mathrm{th}$ or $t_\mathrm{th}$ resulted in a decrease in the number of planes detected. It is interesting to note that $\theta_\mathrm{th} = 6^\circ$ in \prettyref{fig:p2theta_p2t}a and $t_\mathrm{th} = 20^\circ$ in \prettyref{fig:p2theta_p2t}b, the configuration we judged to yield the best results, are turning points: before these points, the number of planes detected decreases quickly; after these points, the number of planes detected decreases much more slowly. As we know, small changes in the configuration of $\theta_\mathrm{th}$ and $t_\mathrm{th}$ should not greatly influence the number of planes detected, so the point cloud may be over-segmented before these turning points. Thus, an analysis of the number of planes detected under different configurations of $\theta_\mathrm{th}$ and $t_\mathrm{th}$ may help find the configuration that yields the best results.

\section{Conclusion}

In this paper, we have proposed an innovative region-growing-based method for automatically extracting outcrop fractures from 3D point clouds. Two local topographic features of the point cloud, i.e., the local surface normal and curvature, are used to define planar surfaces such as fractures. By their definitions, the local surface normal deviation threshold $\theta_\mathrm{th}$ and the transmission error threshold $t_\mathrm{th}$ are designed to control the growth of the fracture regions; $t_\mathrm{th}$ considers the overall occurrence of the fracture while controlling the region growth so that it is not allowed to grow blindly. The orientations are estimated for each extracted fracture.

We tested the proposed method using a 3D point cloud acquired for a real outcrop at the study site, and the results obtained were compared with data collected by a manual field survey for the same outcrop. The test results showed that unlike the existing automatic or semi-automatic methods, the new algorithm can extract the full extent of every individual fracture automatically and accurately. The comparison between our method and the manual field survey shows that the proposed region-growing-based algorithm can obtain fracture data whose quality is as good as or better than that of the manual field survey, thereby demonstrating the potential utility of our method. The performance test using a set of point cloud data showed that the proposed algorithm is suitable for point cloud datasets whose outcrop area $\le$ 70 m$^2$. The analysis of the number of planes detected under different configurations of $\theta_\mathrm{th}$ and $t_\mathrm{th}$ helped explain the configuration we had judged to yield the best results; such analysis may provide a way to find the configuration that yields the best results.

Further research should focus on improving the proposed method by removing possible non-fracture regions (outlier clusters in the fracture sets) and analyzing the results obtained by the region-growing-based algorithm, such as the relationship between the roughness of the fracture and the weathering condition, fracture type, and orientation, as well as assessing the performance of the proposed method with different rock types and weathering conditions.

\section*{Acknowledgments}

This study was funded by the Chinese National Science and Technology Major Project (2011ZX05009001). We are grateful to Prof. Changjiang Li for valuable comments on earlier drafts.  We would like to thank the editor and two anonymous reviewers for their valuable comments and suggestions, which have improved the paper.


\bibliography{\jobname.bib}

\end{document}